\documentstyle{article}
\begin{document}
\section{Introduction}
\label{intro}
\subsection{The philological problem}
In the tradition of ancient Greek texts it sometimes happens that, due to a
number of different reasons a text is credited incorrectly to a certain author.
It is the---sometimes very difficult---task of classical philology to detect
these errorneous assignments and, if possible, to correct them. The methods
used for this task and the results achieved by them are often subject of
strong criticism either because the methods are considered inadequate
or because the results are doubted for some reason.

One of the methods commonly used consists in searching for special stylistic
properties of a given text and comparing them with the personal style of
possible Greek authors of the text in question. By
these means researchers aim at matching properties of the text with the
characteristics of a given writer in order to declare him the author for whom
they have
searched, conversely, or at finding enough contradictions between the text's
and the writer's stylistic properties to be able to negate the authorship of 
the writer.

One special technique is called ``stylometry''. It makes use of several methods
from mathematical statistics. More specifically, it uses statistical tests
to reject a proposed hypothesis by arguing with quantitative data gained
from the text and writer's work.

We want to use stylometry to investigate the authenticity of {\it Rhesus} by
the attic dramatist {\sc Euripides}. This is still an open research topic
in classical philology. Our idea is to test whether the distribution (in the
sense of mathematical statistics) of word categories in the text of
{\it Rhesus} differs significantly from the distribution in the whole work of
{\sc Euripides}.

To do this, we formulate an ``axiom of style'': a typical distribution can be
assigned to every author, and any text written by him follows this distribution
up to a deviation which is statistically irrelevant.

Experiments have shown that two different authors differ in their very own way
to distribute word categories in their text. They consequently can be 
distinguished by computing a distribution of word categories that is typical of
their works. Furthermore, the typical distribution for a single author is
especially influenced by two facts:
\begin{itemize}
\item the literary type of the text
\item the author's age and changes in writing style throughout his lifetime
\end{itemize}

From this observation it follows that there is no typical distribution
{\sc Euripides's} entire works, but only for texts from several limited
temporal periods of his total production time.

\subsection{An idea for implementing a solution}
Therefore, if we want to use this method of stylometry to test {\it Rhesus}
against the rest of {\sc Euripdes'} dramatic texts, we have to try and find
the different stylistic periods in his work and then to compare the
distribution found in {\it Rhesus} with the distribution during each period.

For the Greek text we use the edition according to the{\it Thesaurus Linguae
Graecae} developed by the University of California at Irvine. So all desired
texts are available in digital form. To count the categories in each text we
have implemented a so called ``Part of Speech Tagger'' for attic Greek which
can be adopted to any other language.

\section{Part of Speech Tagging}
\subsection{Mathematical Foundations}
In order to count categories of any text, we need to solve the problem of
assigning a sequence of categories to a given sequence of words, i.e.\ to the
sequence of words forming the text actually under investigation.

Using a part of speech tagger, for any given sequence $W:=(w_1,\dots,w_K)$ one
can estimate the probability of all possible sequences $C:=(c_1,\dots,c_K)$ of
categories for $W$. In part of speech tagging a special sequence $C$ is a
solution for $W$ if and only if it maximizes the probability
\begin{eqnarray}
\nonumber & P\left(C_1 \dots C_K|w_1 \dots w_K\right)&=\\
\label{approx}=& \displaystyle \frac{P\left(w_1 \dots w_K|C_1 \dots C_K\right)\cdot P\left(C_1 \dots C_K\right)}{P\left(w_1 \dots w_K\right)}.
\end{eqnarray}
This says that we search for the sequence $(c_1,\dots,c_K)$ that it most
probable when the sequence of words $(w_1,\dots,w_K)$ is considered. In other
words, (\ref{approx}) is an optimal approxmation to the grammatically correct
sequence of categories valid for $W$.

Due to the enormous amount of data neccessary for the computation of the
probabilities involved in the term described above, it is impossible to
compute a solution directly via this formula.

But to get this complexity under control, we may assume that
\begin{enumerate}
\item for any $i$ in $(c_1,\dots,c_K)$ only the $n+1$ next categories are
dependent of each other, which means that
\begin{equation}
\label{cat} P(c_i|c_{i-1}\dots c_1)\approx P(c_i|c_{i-1}\dots c_{i-n})
\end{equation}
\item for any $i$ in $(w_1,\dots,w_K)$, the category $c_i$ assigned to $w_i$
is not dependent of the preceeding and succeeding categories. So we have
\begin{eqnarray}
\nonumber & P(w_1\dots w_i\dots w_K|c_1\dots c_K)&\approx\\
\nonumber \approx & P(w_1\dots w_{i-1}|c_1\dots c_{i-1})\cdot\\
\nonumber &\cdot P(w_i|c_i)\cdot\\
\nonumber &\cdot P(w_{i+1}\dots w_K|c_{i+1}\dots c_K)&=\\
\label{word} =&\prod_{1\leq i\leq K}P(w_i|c_i)
\end{eqnarray}
\end{enumerate}
Using (\ref{cat}) and (\ref{word}), we now have found an approximation
for $P\left(w_1 \dots w_K|C_1 \dots C_K\right)$:
\begin{eqnarray}
\nonumber & P\left(w_1\dots w_K|C_1\dots C_K\right) &\approx\\
\nonumber = &\prod_{1\leq i\leq K}P\left(C_i|C_{i-1}\dots C_{i-n}\right)\cdot\\
\label{hmm} &\cdot P\left(w_i|C_i\right)
\end{eqnarray}
As it is known from probability theory, (\ref{hmm}) is the defining equation
for a Hidden Markov Model (HMM). From the viewpoint of automata theory, one
can understand a HMM as a stochastic finite state automaton (FSA). For HMMs
there exist algorithms for an efficient solution for (\ref{hmm}). All data we
need, are the following two tables:
\begin{itemize}
\item one table for $P(c_i|c_{i-1}\dots c_{i-n})$ ({\it n-gram probabilities})
\item and one for $P(w_i|c_i)$ (lexical probabilities)
\end{itemize}
It takes a long time to achieve this, because there is no Greek lexicon anywhere that contains the lexical probability of each of its entries; and, because the $n$-gram
probabilities are unknown for any $n$.

To get these neccessary probabilities, we have tagged manually three
texts out of the corpus of {\sc Euripides}, namely {\it Medea} (431),
{\it Electra} (413), and {\it Orestes} (408), in order to cover the entire 
production time of {\sc Euripides} still available to us. Therefore, including
one early, one out of his middle period, and one late tragedy, we have been
able to compute trigram-probabilities for {\sc Euripides} from a sample that
is large enough to get reliable estimations, and that on the average should be
typical for the overall style {\sc Euripides} used in his plays.

For the evalution of lexical probabilities we used the manually tagged texts
to create a small electronic lexicon of Greek used by {\sc Euripides}. This
lexicon is annotated with the lexical probability for each entry. Every
possible category is considered. Because of this, each entry in the lexicon is
stated in list form containing as the first element the Greek word
and, as succeeding elements, all possible categories annotated with the 
adequate lexical probability.

As this lexicon does not contain the entire vocabulary appearing in
{\sc Euripides's} texts, we certainly have to address the problem of unknown
words in the tagging process, if the whole work of {\sc Euripides} is to be
tagged for statistical evaluation. For unknown words no lexical probability
can be found in the lexicon. So how can one compute a solution for
$(w_1,\dots,w_K)$, if any word in this sequence is unknown? On the other hand,
{\sc Euripides} uses many words only once or twice. Thus many frequencies, and
therefore lexical probabilities, are very low. This observation is equally true
for the trigram probabilities.

Common part of speech taggers for modern languages such as English use about
two million words to compute trigram and lexical probabilities. As
{\sc Euripides'} works have a total of only ca.\ 170,000 words, the technique
described above to solve the problem of sparse data cannot be used in our case.

Under these conditions we still have to develop and implement some different
techniques in order to work with sparse data. Surprisingly, when thinking about
the theory of tagging as outlined so far, it is a remarkable fact that no
linguistic, grammatical or syntactic information is used for tagging other 
than the two tables of probabilities. However, when reading text in any
language every human makes immense use of this kind of information in the
process of syntactically and semantically analyzing the sentences one after
another. So a part of speech tagger should also employ the additional
information given by the linguistic structure of any well-formed sequence of
words (This can be a sentence, but also a sequence of sentences).

\section{Tagging with feature value pairs}
The first step in this direction has been taken by Andr\'e Kempe
(\cite{kempe2}): Words do not only carry information about their own category,
but also about gender, number, case, person, etc. This linguistic information
plays an important role in the solution to our category problem. Therefore, if 
we could devise a tagging algorithm that uses this information, we certainly
could improve the tagger's precision.

``Part of Speech Tagging'' is defined as a process that assigns tags denoting a
word's part of speech to the considered word. Up to now, the notion of ``part
of speech'' has been equal to ``word category''. From now on, we do not simply
consider tags which do not only represent the word's category, but also tags
which encode a list of so called feature value pairs. These pairs represent all
linguistic and syntactic information contained in a given word.

As an example, consider the word
$$
\pi\alpha\iota\delta\epsilon\acute{\upsilon}o\mu\epsilon\nu.
$$
It contains the following information:
\begin{itemize}
\item first\ person
\item plural
\item indicative
\item present tense
\item active
\end{itemize}
All of these linguistic facts are coded in a tag appropriate for this word.

Formally we write a tag as
\begin{equation}
t_i:=\bigoplus_{k=1}^{l_i} f_{i,k},\quad \mbox{($l_i$ is the number of
feature value pairs of $t_i$)}
\end{equation}
If $(t_1\dots t_i\dots t_n)$ is a sequence of such tags, the probability
$P(t_i|t_{i-1}\,t_{i-2})$ is computed in the following way:
\begin{eqnarray}
\nonumber P(t_i|t_{i-1}\,t_{i-2})&=&P\left(\bigoplus_{k=1}^{l_i}
f_{i,k}|t_{i-1}\,t_{i-2}\right)\\
\nonumber &=& \frac{\displaystyle P\left(\bigoplus_{k=1}^{l_i}
f_{i,k}\wedge t_{i-1}\,t_{i-2}\right)}{P(t_{i-1}\,t_{i-2})}\\
\nonumber &=& P\left(f_{i,1}|t_{i-1}\,t_{i-2}\right)\cdot\\
&&\cdot \prod_{k=2}^{l_i}P\left(f_{i,k}|\bigoplus_{k=1}^{l_i-1}
f_{i,k} \wedge t_{i-1}\,t_{i-2}\right)
\end{eqnarray}
For lexical probabilities we now write:
\begin{equation}
P\left(w_i|\bigoplus_{k=1}^{l_i} f_{i,k}\right)
\end{equation}
There are two major advantages to this approach. First of all the sparse
data in the case of trigram probablities are more dense now. In addition,
we can save a lot of storage space and execution time during
tagging. For further information, the reader is referred to \cite{kempe2}.

\section{Tagging with morphological analysis}
Kempe's idea is very well suited for improving the tagger's precision in the
case of trigram probabilities. Unfortunately there are many inflected forms
in Greek which induce low frequencies for any lexical entry being an
inflection. Feature value pairs do not provide a solution to this problem.
Therefore, another idea is to be found, when the frequencies of lexicon entries
are to be increased.

Analogously to Kempe's approach additional information should be easily
retrievable from the words in the text to be tagged. But with this
consideration in mind, a straightforward solution can be found: The inflections
themselves carry all information neccessary for determining which word form
they represent. Thus, we only need to split every word into its three defining
parts:
$$
\mbox{prefix -- stem -- suffix}
$$
The prefix appears in the Greek past tenses only. It must be split from the stem,
but carries no information not already contained in stem and suffix. This idea leads to a change in the representation of our lexicon. As the stem is the only
 characteristic part of a word (prefix and suffix are both exchangeable by the
rules of inflection which the Greek grammar defines), our idea is to store the
word's stem alone in the lexicon. Prefix and suffix are then to be stored in
two different tables. 

All words that are not inflections, but have categories such as preposition or
particle,  are stored in the lexicon unchanged as full forms. Thus, in this
case, there is no difference between tagging with or without morphological
analysis.

The fact that in most cases lexicon entries are stems only -- except for
full forms -- has a severe impact on the computation of lexical probabilities.
Since words are split into stem and suffix their lexical probabilities cannot
be computed in the same way as before. The lexicon now only contains
the lexical probability of the stems that can be valid for a given
word. On the other hand, as shown above, an additional table is required which
contains the lexical probabilites of all suffixes. By table lookup we can
determine the lexical probabilities of the suffixes valid for a given word.
Consequently, given the word $w$ with stem $s$ and suffix $u$ ($w=su$) we have
two probabilities now:
\begin{enumerate}
\item $P(t_i|\mbox{stem}=s)$, where $1\leq i\leq n_s$ and $n_s$ is the number
of tags for stem $s$, and
\item $P(t_j|\mbox{suffix}=u)$, where $1\leq j\leq n_u$ and $n_u$ is the
number of tags for suffix $u$.
\end{enumerate}
Assuming independence of these two probabilities, we write:
\begin{eqnarray}
\nonumber & P(t_i|\mbox{stem}=s)\cdot P(t_j|\mbox{suffix}=u) & =\\
\nonumber =& P(t_i|\mbox{stem}=s \wedge \mbox{suffix}=u) & =\\
= & P(t_i|w)\approx P(w|t_i)
\end{eqnarray}
The last step is due to Bayes' rule\footnote{\cite{kempe1} shows that in the
case of part of speech tagging both formulas are almost equivalent.}. Of
course, step 2 is valid if and only if suffix and stem form a correct Greek
word. For example,
$$
\pi\alpha\delta\epsilon\acute{\upsilon}-\sigma\alpha\nu\tau o\varsigma
$$
is correct, while
$$
\pi\alpha\delta\epsilon\acute{\upsilon}\sigma\alpha\nu\tau-o\varsigma
$$
is not. But in both cases the suffixes are correct. Therefore, during the
computation of lexical probabilities there must be a check to determine
whether the computed probability is valid at all.

There are some advantages with this approach concerning the computation of
lexical probabilities:
\begin{itemize}
\item Frequencies for stems and suffixes are higher when considered separately.
Thus, tagging precision can be further increased.
\item In the case of unknown words, enough information can be extracted from
the suffix to significantly limit the number of tags possible for the given 
word compared to the total number of tags. Taggers commonly consider all tags
possible for any unknown word. Thus, morphological analysis vastly increases
the precision of lexical probabilities during tagging.
\item No lexicon for full forms is neccessary. For highly inflected langugages,
this means an enormous decrease of resources needed for lexicon storage
and access.
\end{itemize}
\section{Implementation}
For our task of tagging the corpus of {\sc Euripides} we implemented the
morphological analysis using regular expressions which describe almost all of
the suffixes of the Greek language and the associated tags. On the basis of
these regular expressions we implemented a parser that precedes the tagging
process and annotates all words with their possible tags and respective lexical
probabilities. This parser is implemented in C. As everything, and especially
all data, had to be built from scratch, this seemed to be the most appropriate
way to reach our goal. But morphological analysis could be achieved in quite a
different way, too.

Our complete tagging system was trained with ca.\ 17,000 words (see above) and
works with an average accuracy of about 96,6\%. The lexicon has about 5,900
entries. Due to different accentuation marks possible for the same stem, some
of these entries actually occur twice. So the number of different Greek words
stored in the lexicon is even lower.

Another point to be made here is the fact that many other modern languages
form words by inflection and suffixes giving the stem a special meaning and
denoting special categories or tags in every case.
From this point of view it becomes clear that morphological analysis is a
powerful tool for part of speech tagging not only for ancient Greek. Here are
some examples for suffixes in English, German, and Italian:
\vskip\baselineskip
\begin{tabular}{lll}
{\bf German} & {\bf English} & {\bf Italian}\\
-ig & -ness & -ista\\
-ung & -ize & -tore\\
-heit & -ate & -trice\\
-en & -tion & -t\'a\\
-er & -ing & -ndo\\
-el & -ed & -bile\\
-erin & -ly & -mente\\
\end{tabular}

\section{Application to the corpus of {\sc Euripides}}
\subsection{Statistics}
First we used our tagger to create tables which count the number of words in
each category for all eighteen works of {\sc Euripides} that are not lost
during the tradition of the ancient texts including the critical {\it Rhesus}.

On the basis of these data we try to give answers on the questions posed in
section \ref{intro}. To measure the extent to which a certain work deviates
from the distribution of several others, we use the $\chi^2$-test with two
classes:
\begin{equation}
\chi^2:=\sum_{1\leq i\leq2}\frac{(M_{i,j}-p_{i,j})^2}{p_{i,j}}
\end{equation}
$p_{i,j}$ is the probability for category $j$ to appear ($i=1$) or to not
appear ($i=2$), while $M_{i,j}$ is the number of words of category $j$ ($i=1)$
or the number of words of category different from $j$ ($i=2$).

For every text tested we compute the sum of all significant deviations
$\alpha_i$ and measure their distance from the common mean value $\mu$ in
respect to the common standard deviation $\sigma$ by
\begin{equation}
\rho_i:=\frac{\alpha_i-\mu}{\sigma}
\end{equation}
$\mu$ and $\sigma$ are estimated by Maximum Likelyhood Estimation.
$\rho_i$ is considered significant if and only if $\rho_i\geq 2$.
\subsection{Results}
In the corpus of {\sc Euripides} there are several works whose date of creation
is supposed to be between 413 and 408, {\it Helena}, {\it Ion}, {\it Iphigenia
Taurica}, and {\it Phoenissae}. In a first series of tests we searched for
significant deviations between these four works. None could be found. However, 
the value for deviation became larger, though still unsignificant, when taking 
into account {\it Electra}, too. But {\it Orestes}, on the other hand,
seems to match smoothly into the pattern defined by the other four texts.
From these observations we conclude that {\it Electra} could mark the
transition from the middle to the late period of {\sc Euripides'} work. But we
have not discovered any new facts which could make clearer the time of writing 
of the other four works considered in these tests.

Next we tried to find a transition from the early to the middle period. To do
this, we tested {\it Alcestis}, {\it Medea}, and {\it Heraclidae} against
several works from the middle and all from the late period. All proved to
deviate significantly. So, by means of this test, they are characterized as
early works. For {\it Hippolytus} in the first attempt we could not find a
significant deviation. But, after exchanging the possibly
spurious {\it Rhesus} with {\it Helena} out of the late period and {\it Hecuba}
with {\it Electra}, we received significant values for {\it Hippolytus} as
well. Therefore, we believe that {\it Hippolytus} and {\it Hecuba} mark the 
transition from the early to the middle period. This stylistic characterization
of {\sc Euripides'} work coincides precisely with its chronological order.

In a final series of tests we wanted to find out something about the behaviour
of {\it Rhesus}. First we found a significant deviation from the early period.

As an example, we show a table containing the evaluation of the test which
compares {\it Rhesos} and the early plays listed below (The first column
contains the names of the different word categories considered. The last row
shows the values for $\rho$. All other numbers are $\chi^2$-values.):
{\tabcolsep=1.00pt \footnotesize \vskip\baselineskip
\begin{center}\sf
\begin{tabular}{lcccccc}
& Alcestis &  Medea &  Heraclidae &  Hippolytos &  Hecuba & Rhesus\\
adjk & 0.0909& 1.79& 1.28& 0.911& 0.0500& 0.922\\
adjp & 21.7& 15.7& 9.48& 0.846& 6.57& 0.188\\
adjs & 3.37& 5.77& 0.270& 0.647& 4.31& 3.67\\
adva & 0.117& 2.60& 1.63& 4.98& 0.273& 0.0796\\
advs & 0.862& 0.00389& 0.918& 0.101& 0.171& 4.76\\
arti & 0.383& 4.97& 4.25& 4.30& 2.72& 11.1\\
depn & 0.567& 0.0849& 20.1& 1.52& 1.59& 7.17\\
idpn & 0.0684& 0.00189& 0.154& 0.0707& 0.957& 0.558\\
intj & 4.98& 0.176& 0.122& 13.9& 3.93& 21.3\\
irpn & 2.48& 0.0970& 0.600& 0.174& 1.11& 0.292\\
konj & 1.22& 2.70& 0.165& 3.63& 2.96& 0.0217\\
name & 8.06& 31.9& 2.07& 11.9& 14.7& 69.0\\
nega & 3.66& 0.257& 0.171& 2.10& 4.65& 1.29\\
nume & 0.0149& 0.0132& 0.000453& 5.49& 1.48& 1.40\\
parl & 3.78& 0.629& 7.44& 3.58& 7.69& 10.8\\
part & 4.67& 5.51& 0.278& 7.47& 0.0921& 13.3\\
pepn & 3.94& 0.801& 0.00385& 0.357& 0.193& 3.76\\
popn & 0.252& 2.59& 2.84& 0.371& 1.18& 7.18\\
prae & 0.459& 5.07& 2.07& 0.233& 0.0918& 8.72\\
repn & 0.110& 0.346& 0.482& 0.00170& 0.00852& 1.51\\
rlpn & 0.0156& 2.56& 3.01& 0.165& 0.987& 0.775\\
subs & 1.96& 1.04& 11.2& 5.11& 3.92& 13.3\\
verf & 5.17& 0.168& 0.0111& 0.00109& 0.853& 0.542\\
veri & 0.693& 0.00435& 5.10& 2.26& 5.02& 1.20\\\hline
&$- 0.870$&$- 0.124$&$- 0.124$&$- 0.124$&$- 0.870$& {\bf 2.10}\\
\end{tabular}\end{center}\vskip\baselineskip}
The value of $2.10$ for {\it Rhesus} indicates that there are significantly
many deviations of the $\chi^2$-values for {\it Rhesus} compared with the
overall mean values. This says that in the text of {\it Rhesus} too many
word categories appear in too small or too large a number in order to be still
characteristic for the typical numbers of appearances in the other texts
included in this test. From this result we conclude that {\it Rhesus} has not
been written at the same time as the other five plays.
 
We got the same result when testing the middle period only and both early and
middle periods.

Only when compared with the late period, does {\it Rhesus} show no significant
deviation at all.

We therefore conclude that, when using our ``axiom of style'' as a criterion,
the hypothesis of {\it Rhesus} being a early or middle-period-work by
{\sc Euripides} can be rejected. However {\sc Euripides} could still have
written {\it Rhesus} in his late period or even before 438. Unfortunately, not 
enough text material from before 438 exists to test {\it Rhesus}
by means of statistical techniques. Nevertheless, the deviations found for
{\it Rhesus} point out to the fact that the time of its writing is
strongly restricted to either the very early or very late lifetime of
{\sc Euripides}. The question of authenticity, however, still remains open.

\raggedright 
\end{document}